\documentclass{article}

\usepackage{arxiv}

\usepackage[utf8]{inputenc} 
\usepackage[T1]{fontenc}    
\usepackage{hyperref}       
\usepackage{url}            
\usepackage{booktabs}       
\usepackage{amssymb,amsfonts}       
\usepackage{nicefrac}       
\usepackage{microtype}      
\usepackage{lipsum}
\usepackage{graphicx}
\usepackage{multirow}
\usepackage{algorithmic}
\usepackage{algorithm}
\graphicspath{ {./} }

\title{Determining Sequence of Image Processing Technique (IPT) to Detect Adversarial Attacks}

\author{
 Kishor Datta Gupta \\
  University of Memphis\\
  Tennessee, USA \\
  \texttt{kgupta1@memphis.edu} \\
  \And
 Zahid Akhtar \\
   University of Memphis\\
  Tennessee, USA \\
  \texttt{zmomin@memphis.edu} \\
   \And
 Dipankar Dasgupta \\
   University of Memphis\\
  Tennessee, USA \\
  \texttt{ddasgupt@memphis.edu} \\
}

\begin{document}
\maketitle
\begin{abstract}
Developing secure machine learning models from adversarial examples is challenging as various methods are continually being developed to generate adversarial attacks. In this work, we propose an evolutionary approach to automatically determine Image Processing Techniques Sequence (IPTS) for detecting malicious inputs. Accordingly, we first used diverse set of attack methods including adaptive attack methods (on our defense) to generate adversarial samples from the clean dataset.  A detection framework based on a genetic algorithm (GA) is developed to find the optimal IPTS, where the optimality is estimated by different fitness measures such as Euclidean distance, entropy loss, average histogram, local binary pattern and loss functions. The ‘image difference’ between the original and processed images is used to extract the features, which are then fed to a classification scheme in order to determine whether the input sample is adversarial or clean. This paper described our methodology and performed experiments using multiple data-sets tested with several adversarial attack. For each attack-type and dataset, it generates unique IPTS. A set of IPTS selected dynamically in testing time which works as a filter for the adversarial attack. Our empirical experiments exhibited promising results indicating the approach can efficiently used as processing for any AI model.
\end{abstract}

\section{Introduction}

\begin{figure}[hbt!]
\centering
  \includegraphics[scale=0.7]{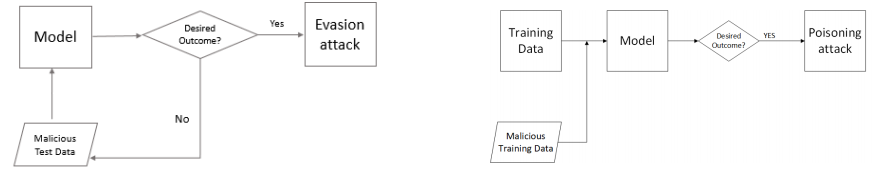}
  \caption{Evasion attack and poisoning attack generation \cite{ibitoye2019threat}  }
  \label{fig:eva}
\end{figure}

Though machine learning models in recent years have shown success in many applications, they are vulnerable to adversarial attacks. According to NIST\cite{tabassi2019taxonomy} definition of adversarial machine learning (AML) can be the manipulation of training data, ML model architecture, or manipulating testing data in a way that will result in wrong output from ML system. Attacks on the ML system in the training phase is known as poisoning or causative attacks. Whereas in the testing phase, they are called evasion or exploratory attacks. We only consider evasive attacks in this study. In evasive attacks, the testing samples are altered in such a way that they are wrongly classified to different targeted or non-targeted class as in figure \ref{fig:eva}, we can see that malicious training data was given to the model with training data until the desired outcome start to happen.. Studies have shown that to classify an object in an image, deep neural networks (DNNs) usually do not identify the object's macro level features, e.g., the difference between a plane and a car does not depend on the background (i.e., sky or ground, or which one has wings or large windows). Rather, it depends on color pixel values at certain positions \cite{biggio2013evasion}. In 2014, Szegedy et al. \cite{szegedy2013intriguing} discovered that DNNs have some intriguing properties, e.g., very slight random perturb distribution can cause an image to be misclassified. Goodfellow et al. \cite{goodfellow2014explaining} later that year showed that such perturbed distribution is not random; they follow specific characteristics. Attackers need to generate adversarial samples with effective traits.
\begin{table}[hbt!]
\centering
\begin{tabular}{|c|c|c|c|}
\hline
Acronym & FullMeaning & Acronym & Full Meaning \\ \hline
BS & Bilateral smoothing & BPDA & \begin{tabular}[c]{@{}c@{}}BackPass Differential\\ Approximation\end{tabular} \\ \hline
AS & Adaptive Smoothing & PGD & \begin{tabular}[c]{@{}c@{}}Projected Gradient\\   disent\end{tabular} \\ \hline
AN & Additive Noise & BIM & \begin{tabular}[c]{@{}c@{}}Basic Iterative\\   method\end{tabular} \\ \hline
FGSM & \begin{tabular}[c]{@{}c@{}}First Gradient Sign\\   Method\end{tabular} & MBIM & \begin{tabular}[c]{@{}c@{}}Momentum Basic\\   Iterative Method\end{tabular} \\ \hline
JSMA & \begin{tabular}[c]{@{}c@{}}Jacob Saliency Map\\   Method\end{tabular} & SIPTS & \begin{tabular}[c]{@{}c@{}}Set of Image\\   Processing Technique \\Sequence\end{tabular} \\ \hline
DF & Deep Fool method & DI & \begin{tabular}[c]{@{}c@{}}Pixel Difference\\   between before and\\ after IPTS applied\end{tabular} \\ \hline
HSJ & \begin{tabular}[c]{@{}c@{}}HopSkipJump\end{tabular} & ED & Euclidean distance \\ \hline
MLM & \begin{tabular}[c]{@{}c@{}}Machine Learning\\   Model\end{tabular} & LBP & Local binary pattern \\ \hline
IPT & \begin{tabular}[c]{@{}c@{}}Image Processing\\   Techniques\end{tabular} & PDE & \begin{tabular}[c]{@{}c@{}}Probability Density\\   Equation\end{tabular} \\ \hline
IPTS & \begin{tabular}[c]{@{}c@{}}Image Processing\\   Technique Sequence\end{tabular} & TN & Thining \\ \hline
PX & \begin{tabular}[c]{@{}c@{}}Pixellate \end{tabular} & GS & Grey-scaled \\ \hline
\end{tabular}
\caption{Acronym used in this paper}
\label{tab:notation}
\end{table}

\begin{figure}
\centering
  \includegraphics[scale=0.5]{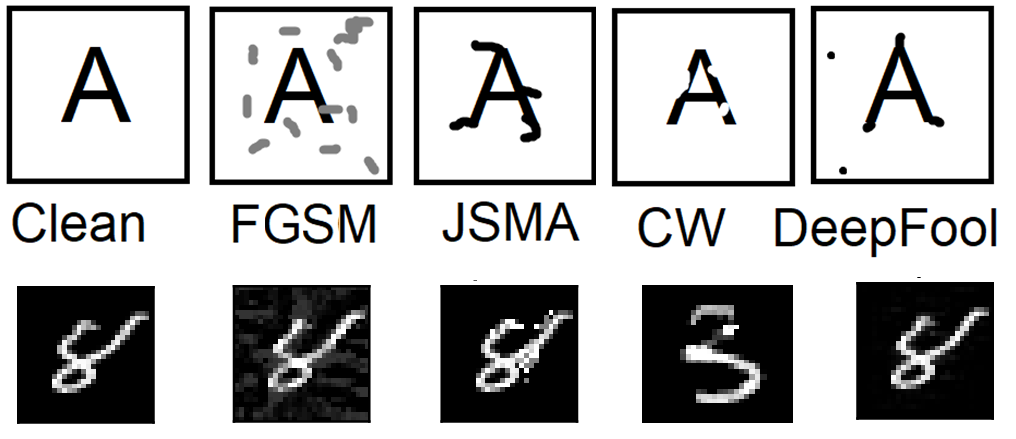}
  \caption{First rows show how different types of attack change a clean image (for visual purpose, the effects are exaggerated. This row is created from observation and not real sample). The second rows are actual corresponding real attacks on the MNIST dataset. Where 8 recognized as a different label. }
  \label{fig:image}
\end{figure}

\begin{figure}
\centering
  \includegraphics[scale=0.45]{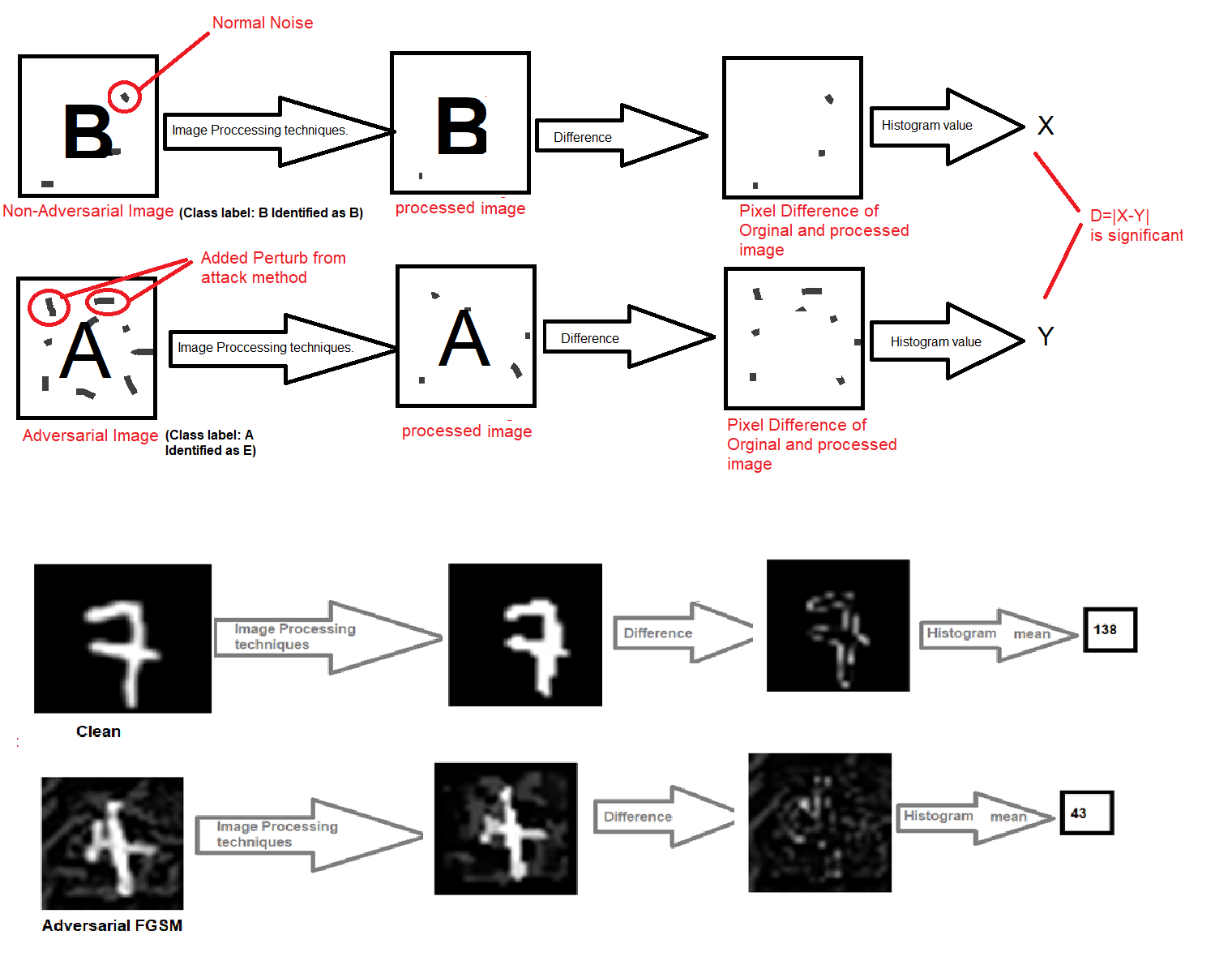}
  \caption{In the top side, Clean (class A) vs Adversarial (B which classified as other class) images differences after IPTS applied, Histogram calculation on the difference, In bottom, MNIST with FGSM example has shown}
  \label{fig:imagec}
\end{figure}

\begin{table*}[]
\centering
\begin{tabular}{|l|l|l|l|}
\hline
\multicolumn{1}{|c|}{\textbf{\begin{tabular}[c]{@{}c@{}}Data\\   Mnaipulation attack\end{tabular}}} & \multicolumn{1}{c|}{\textbf{\begin{tabular}[c]{@{}c@{}}Sequence  for \\ Dataset (MNIST)\end{tabular}}} & \multicolumn{1}{c|}{\textbf{\begin{tabular}[c]{@{}c@{}}Sequence for Dataset\\ (CIFAR)\end{tabular}}} & \multicolumn{1}{c|}{\textbf{Reason}} \\ \hline
FSGM & BS+BS & Greysale+AN+AN & \begin{tabular}[c]{@{}l@{}}perturbation which added in image is not an \\edge, so edge-preserving algorithm  remove these, \\so differences of before after\\ adversarial image inputs are \\higher than non=adverse image\end{tabular} \\ \hline
JSMA & TN+BS & PX+greyscale+BS & \begin{tabular}[c]{@{}l@{}}JSMA extend some edge which\\   reduced when we do thining\end{tabular} \\ \hline
CW & \begin{tabular}[c]{@{}l@{}}TN+TN+TN or\\ GS+GS+GS+TBS\end{tabular} & \begin{tabular}[c]{@{}l@{}}GS+greyscale +GS+TBS\\+TN+TN\end{tabular} & \begin{tabular}[c]{@{}l@{}}In CW object edge getting thining\\ and blur,so applying more\\  blur and thining algorithm diffence will applify\end{tabular} \\ \hline
DeepFool & GS+AN+GS+AN & GreySale+AN+AN+AN & \begin{tabular}[c]{@{}l@{}}Deepfool create few pixel\\ arround border to effect the model,using \\ additive noise boost this effect and\\ we can have a difference\end{tabular} \\ \hline
\end{tabular}
\caption{Different Image processing techniques for different adversarial attack type}
\label{tab:tab2}
\end{table*}


\begin{table}[]
\centering
\begin{tabular}{|c|c|c|c|c|c|c|c|}
\hline
\multirow{4}{*}{Threat Model} & \multicolumn{2}{c|}{\textbf{ATTACK TYPE}}                                                             & \textbf{JSMA} & \textbf{FGSM} & \textbf{DF} & \textbf{HSJ} & \textbf{BPDA} \\ \cline{2-8} 
                                                                                   &
                                                                                   \multirow{2}{*}{Adversarial  Knowledge}    & WhiteBox       &*            &*            &*          &              &               \\ \cline{3-8} 
                                                                                   &                                                                                      & BlackBox       &               &               &             &*           &               \\ \cline{2-8} 
                                                                                   & \multirow{2}{*}{Adversarial  Specificity} & Target ted      &*            &               &             &*           &               \\ \cline{3-8} 
                                                                                   &                                                                                      & Non-Target ted  &               &*            &*          &              &               \\ \cline{2-8} 
                                                                                   & \multirow{2}{*}{Attack   Frequency}      & One time       &               &*            &             &              &               \\ \cline{3-8} 
                                                                                   &                                                                                      & Iterative      &*            &               &             &              &               \\ \cline{2-8} 
                                                                                   & \multirow{2}{*}{Adversarial Methodology}   & Gradient based &*            &*            &*          &              &               \\ \cline{3-8} 
                                                                                   &                                                                                      & Gradient free  &               &               &             &*           &               \\ \cline{2-8} 
                                                                                   & Adaptive Attack                                                                      &                &               &               &             &              &*            \\ \hline
\multirow{3}{*}{BenchMark}    & \multirow{3}{*}{Dataset}                                                             & MNIST          &*            &*            &*          &              &               \\ \cline{3-8} 
                                                                                   &                                                                                      & CIFAR          &*            &*            &*          &*           &               \\ \cline{3-8} 
                                                                                   &                                                                                      & IMAGENET       &*            &*            &             &              &               \\ \cline{2-8} 
                                                                                   & \multirow{3}{*}{ML Model}                                                            & ResNet         &*            &*            &             &              &               \\ \cline{3-8} 
                                                                                   &                                                                                      & Simple CNN     &               &               &*          &*           &*            \\ \cline{3-8} 
                                                                                   &                                                                                      & VGG            &               &*            &             &              &               \\ \hline
\end{tabular}
\caption{Summary of Threat Models and associated manipulation strategies and dataset used for experimentation.}
\label{tab:ThreatModel}
\end{table}

Generally speaking, adversarial examples are input data that get misclassified by a ML/AI system but not by a human subject. Several defense methods against adversarial examples, such as adversarial training, defensive distillation, etc., have been proposed \cite{xu2017feature}. All have these methods have some drawbacks which we discussed briefly in this paper. Image Prepossessing technique based defense against adversarial input currently considering obsolete as this defense are not effective against all types of attacks and can easily be bypassed by the advanced adaptive attack. In this paper, we proposed a framework which will generate a set of image processing sequence ( which made by several image processing techniques) that can detect the diverse set of adversarial inputs. We also showed that our method works against adaptive attacks as we are using a dynamic selection of set of Image processing technique sequences(SIPTS). As we found that different sequences of IPTS are effective against different attack types, which we can see in table \ref{tab:tab2}. Determining these sequences are hard as for only 4 IPTS (and any IPT can be repeated) then the total possible sequences, considering sequence max length 8, would be $1^4+2^4+3^4+4^4+…+8^8= 24,684,612$. Thus, finding the right sequence from that many possible sequences is a massive task, which here we have attempted to solve using a GA. In this study, several metrics such as histogram of difference image (DI), Euclidean distances (ED), cross-entropy (CE) value, loss function, and probability density value were explored to measure effectiveness of IPTS. Specifically, DI is the pixelwise difference image of original input image and processed input image. Histogram represents the pixel distribution of the image \cite{kannala2012bsif}. Histogram does not concern about shape, size or any attribute rather than color distribution of an image. Histogram is obtained by plotting color range on $x$ axis and number of pixels with specific color on $y$ axis. Average Histogram ($H_{avg}$) is the average value of $y$ axis. Our proposed solution is to obtain an evolutionary algorithm that will determine a set of IPTS sequences $IPTS$ and $H_{avg}$ range. When this $IPTS$ applied to an input image, it will have a $H_{avg}$ value, which within an estimated range is classified as adversarial input. Another feature is Euclidean distances between adversarial and clean DI histograms. In testing time, we compute ED of input image, DI histogram with adversarial DI histograms and Clean DI histogram and see which one has less distance. We also used cross-entropy value (CE) and probability density equation (PDE) cross entropy of adversarial and clean DI histograms. We also calculate loss function of average histogram distribution of adversarial and clean histograms DI. All studied metrics are detailed in section \ref{proposedmethod}. In testing time, we are selecting our set of  Image processing sequences dynamically to answer the obscurity question. This paper outlined our methodology and conducted experiments utilizing varied data-sets examined with diverse adversarial data manipulations. For specific attack types and dataset, it produces unique IPTS. Our empirical experiments presented encouraging outcomes showing the method can efficiently be employed as processing for any machine learning models as it also compliant with cyber security principles. 
\begin{figure}
\centering
  \includegraphics[scale=0.37]{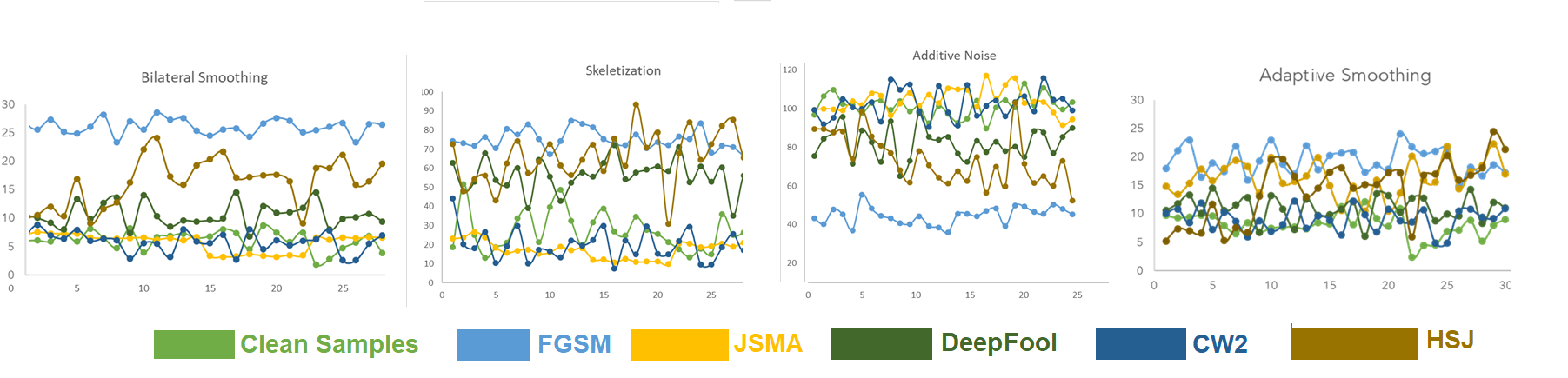}
  \caption{4 IPT applied to 5 types of the adversarial set with a clean sample set for the MNIST dataset. In X-axis, no of each sample shows and Y axis, Before and after effects histogram average has been illustrated. From this small set we can see, different IPT has different effects for different attack types }
  \label{fig:IPT}
\end{figure}

.

In Section \ref{background}, Adversarial attacks, existing adversarial defense works, IPTs, and basic of GA re described. The proposed method with threat model and implementation details are presented in Section \ref{proposedmethod}. Experimental results and discussions are given in Section \ref{experimentsresults} and \ref{resultssectiona}. Due to space restriction, we only provided an experiment procedure with JSMA attacks as other attack types follow a similar procedure. In Section \ref{conclusion}, conclusions are drawn with our limitations and future work plan. 
\section{Background}
\label{background}
In this section, we mentioned related works for adversarial input detections, some background of adversarial attack, the basic concept of Genetic algorithm and image processing techniques.
\subsection{Related work}
In 2017, Grosse et. al. did statistical tests using a complementary approach to identify specific inputs that are adversarial \cite{grosse2017statistical}. Wong et. al. shows convex outer adversarial polytope can be a proven defense \cite{wong2017provable}. Lu et. al. (2017) checks depth map is consistent or not (only for image) to detect adversarial examples \cite{lu2017safetynet}. Metzen et. al. implemented deep neural networks with a small “detector” sub-network which is trained on the binary classification task of distinguishing factual data from data containing adversarial perturbations \cite{metzen2017detecting}. The same year, Madry et. al. (2017) published a paper on adversarial robustness of neural networks through the lens of robust optimization \cite{madry2017towards}. Chen et. al. tried to devise adversarial examples with another guardian neural net distillation as a defense from adversarial attacks\cite{chen2017reabsnet}. In 2018, Wu et. al. developed Highly Confident Near Neighbor (HCNN), a framework that combines confidence information and nearest neighbor search, to reinforce adversarial robustness of a base model\cite{wu2018reinforcing}. Same year, Paudice et. al. applied Anamoly Detection\cite{paudice2018detection} and Zhang et. al. detected adversarial examples by identifying significant pixels for prediction which only work for image \cite{zhang2018detecting,}. Other researchers such as Wang et. al. tried with mutation Testing \cite{wang2018detecting} and Zhao et. al. developed key-based network, a new detection-based defense mechanism to distinguish adversarial examples from normal ones based on error correcting output codes, using the binary code vectors produced by multiple binary classifiers applied to randomly chosen label-sets as signatures to match standard images and reject adversarial examples \cite{zhao2018detecting}. Later that year Liu et. al. tried to use steganalysis\cite{liu2018detection} and Katzir et. al. implemented a filter by constructing euclidean spaces out of the activation values of each of the deep neural network layers with k-nearest neighbor classifiers (k-NN) \cite{katzir2018detecting}. A different notable strategy was taken by researchers Pang et al. They used thresholding approach as the detector to filter out adversarial examples for reliable predictions\cite{pang2018towards}.  For an image classification problem, Tian et. al. did image transformation operations such as rotation and shifting to detect adversarial examples\cite{tian2018detecting} and Xu et. al.\cite{xu2017feature} simply reduced the feature space to protect against adversary. In 2019, Monteiro et al \cite{monteiro2018generalizable} developed inputfiler which is based on Bi-model Decision Mismatch of image. Sumanth Dathathri showed whether prediction behavior is consistent with a set of fingerprints (a data set of NN) named NFP method \cite{dathathri2018detecting}. Same year, Crecchi et. al. used non-linear dimensionality reduction and density estimation techniques \cite{crecchi2019detecting} and Aigrain et. al. tried to use confidence value in CNN\cite{aigrain2019detecting}. Some other notable works in that year are meta-learning based robust detection method to detect new adversarial attacks with limited examples developed by Ma et. al. \cite{ma2019metaadvdet}. Another important and effective work done by  Chen et. al., where they tried to keep the records of query and used KNN to co-relate that with adversarial examples \cite{chen2019stateful}

In summary, defenses against adversarial attacks can be classified in two ways: attack detection and robust recognition model. Detection is a binary classification problem, where input is classified as an adversarial or not. In robust recognition methods, the correct class of an adversarial image is recognized where the attack classed are known prior. This work is focused on detection technique. Well-known robust recognition models are training on adversarial inputs proactively \cite{goodfellow2014explaining}, performing defensive distillation \cite{papernot2016distillation}, training the network with enhanced training data all to create a protection against adversarial example \cite{miyato2016adversarial}. Image histogram-based \cite{prakash2018deflecting} methods are also used to detect adversarial inputs. In \cite{akhtar2018adversarial}, authors proposed an adversarial attack detection scheme based on image quality related features to detect various adversarial attacks. Carlini et. al. \cite{carlini2017adversarial} tested ten defense techniques, by detailed evaluation they showed that pre-processing techniques can be easily bypassed. Also, a brief summary of most attacks and all defense techniques is described by Akhtar et. al. \cite{akhtar2019brief}. While, our proposed method works as a defense (attack detection) technique against adversarial input in testing time.

\subsection{Adversarial attack}

Rauber et. al. \cite{rauber2017foolbox} mentioned three types of adversarial image manipulation attacks, i.e., gradient, score, and decision based. In particular First gradient sign method, which is a gradient based strategies, was proposed by Ian Goodfellow in 2014. Carlini and Wagner (CW) \cite{xu2017feature} uses an iterative attack that constructs adversarial examples by approximately solving the minimization problem. CW is different from gradient-based methods such that it is an optimization-based attack. The Jacobian-based saliency map attack \cite{papernot2016distillation} is another class of adversarial attack for deceiving classification models. A saliency map is an image that shows each pixel's unique quality. The purpose of a saliency map is to simplify and improve the representation of a picture into something more significant and better suitable to analyze. For instance, if a pixel has a high grey level or other unique color quality in a image, that pixel's class will show in the saliency map and in an obvious way. Saliency is a kind of picture segmentation. DeepFool \cite{moosavi2016deepfool}(DF) attacks find the closest distance from the original input to the decision boundary.

\begin{figure*}
\centering
  \includegraphics[scale=0.7]{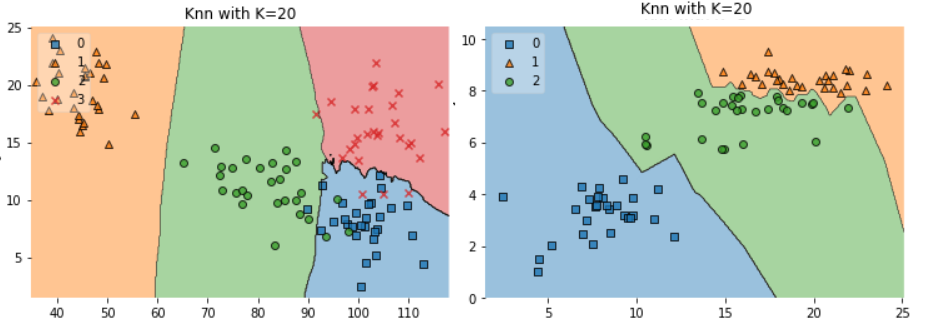}
  \caption{In the right side, clean (0), FGSM(1) and JSMA(2) samples were applied with AS IPT and using KNN\cite{zhang2007ml} with the Histogram average value of DI and euclidean distance value,Similarly in the left side, clean (0), FGSM(1) and JSMA(3) and DF(2) samples were applied with AN + AS IPT s effects. We can see DF, Jsma is overlapping there.}
  \label{fig:imageknn}
\end{figure*}

\subsection{Image Processing Techniques (IPT)}

IPT can be seen as data transformation/conversion techniques, which can be used to modify any image to another image where specific properties of the original image are modified/transformed. We used several IPTs such as adaptive smoothing (AS)\cite{ramponi1996rational}, additive noise(AN), bilateral smoothing (BS), Gaussian blur, sharpen, thickening for example. AS is an edge-preserving smoothing technique\cite{zhiyuan2005robust}. Additive white Gaussian noise (AWGN)\cite{zhao2007image} is a type of basic noise model which adds noise by mimicking the natural process \cite{kevin2000radio}. It adds some extra pixel based on the neighborhood pixel distribution. The BS\cite{zhiyuan2005robust} is another type of smoothing technique based on gaussian functions. Gaussian blur or Gaussian smoothing is a technique to reduce the picture quality.  In decreased quality image, the noise also gets decreased much more than non-noise pixels. Convolving picture with a Gaussian function is a technique to execute this process. Gaussian blur is recognized as a low pass filter as it decreases the image's higher frequency components \cite{waltz1998efficient}. Image processing technique has been used for classification tasks such as face orientation classification \cite{gupta2017robust} or improve image quality \cite{gupta2017robust}.

\begin{figure*}
\centering
  \includegraphics[scale=0.3]{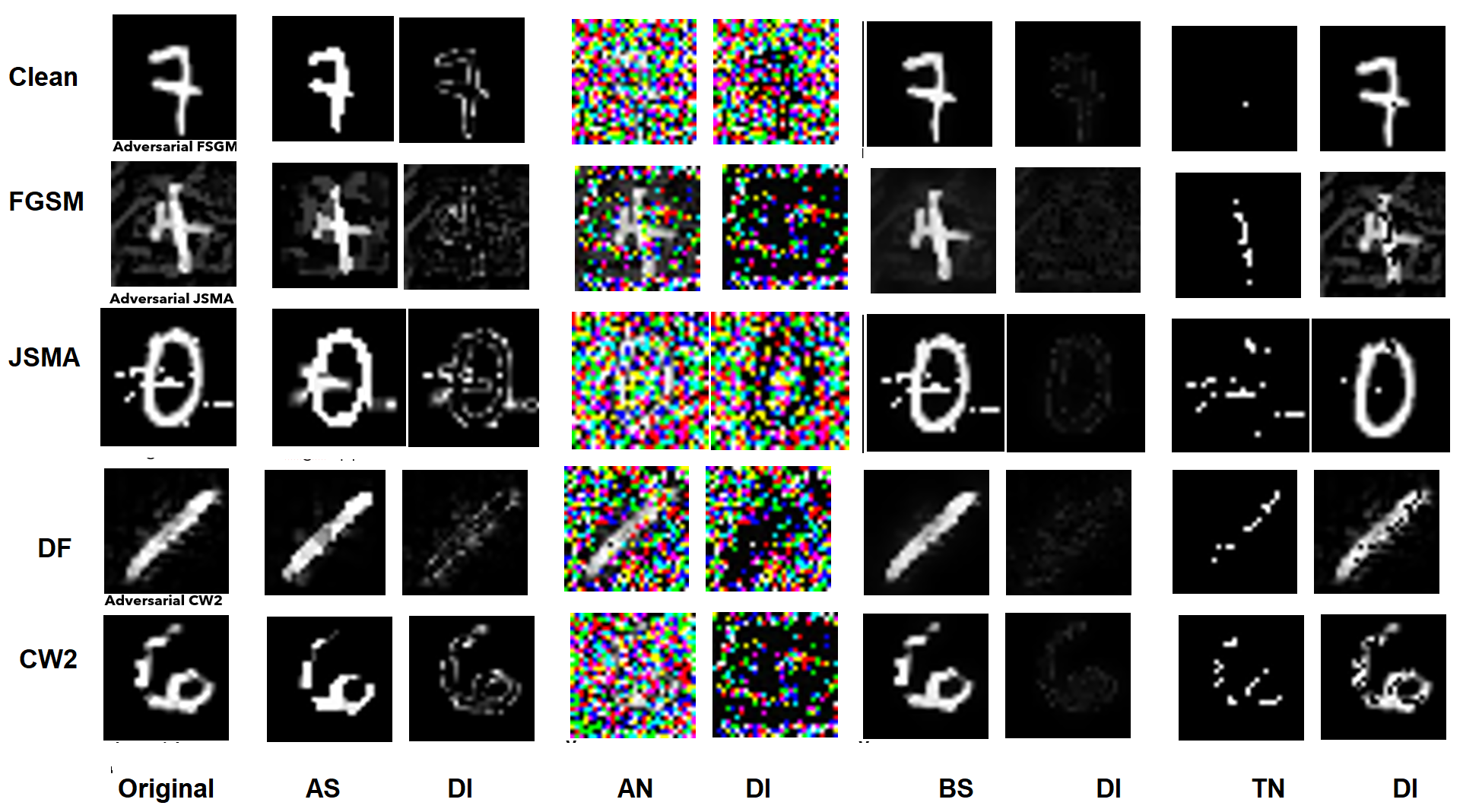}
  \caption{Effects of Several IPT techniques on MNIST dataset, after applying different IPT}
  \label{fig:mnist}
\end{figure*}

\subsection{Genetic Algorithm (GA)}

GA \cite{koza1995survey} is a nature-inspired, population based heuristic search. At first, a set of random solutions is generated from all possible solutions exist in representation space. Then an evaluation of these solutions based on some measures is performed. Based on these measures, the best solutions are crossover and muted to generate a new set of solutions. The same procedure is iterated again and again until a near optimal solution found or termination condition met. During this process, a solution with the best measures is selected as a result. Each problem properties encoded as a gene. The set of genes creates individuals as their solution, a set of individuals are known as population. When two individuals are mixed and create a new individual, it is known as crossover. When a bit of an individual is changed, it is known as a mutation in the GA. GA has been used for Computer Vision (regenerate images \cite{gupta2018genetic}), Computational AI domain (searching recommendations\cite{gupta2018smart}\cite{sadman2020detect},Cyber security Domain ( smart grid management \cite{senagenetic}),and many other domains.

\section{Proposed Framework}
\label{proposedmethod}
In this section, we provided the threat model we are considering while devising our defense, basic hypothesis of our techniques and detailed different steps of our proposed framework.
\subsection{Threat model}
Table \ref{tab:ThreatModel} summaries the threat models we investigated in our work. Here the second column shows different strategies used in threat models, the third column mentions usability tactics and the rest of the columns provide specific image manipulation techniques. Yuan et al. (2018)\cite{yuan2019adversarial} suggested making threat models consist of Adversarial Falsification (False negative, False Positive), white-box, BlackBox, targeted, non- targeted, onetime and iterative attacks. 
Carlini et al.\cite{carlini2019evaluating}, suggested that adversarial attack and defense models need to be tested against a diverse set of attacks. Also, they need to be evaluated against adaptive attacks. Moreover, Tramer et al. \cite{tramer2020adaptive} suggested different themes to evaluate a defense model. Keeping these guidelines in mind, we developed our threat model inclusive of basic, advanced attack and adaptive attack(against our defenses) types. However, we skipped the falsification category as this threat model does not bring much variety. As Carlini et al. \cite{carlini2017adversarial} recommended using at least one gradient-free and one hard-label attack. To address that concern, we also evaluated our proposed method with gradient-free attacks such as local search attack \cite{narodytska2016simple} and hop-skip-jump attack \cite{chen2019hopskipjumpattack}. To compare our defense work against suggested attack types, we tested with basic iterative methods (BIM) \cite{madry2017towards} and their variations (e.g., MBIM\cite{dong2018boosting} and PGD). We selected CW2 and DF attacks as advanced attacks. For testing against an adaptive attack, we used BPDA (Backward Pass Differential Approximation). It is proposed by Athaye et al. \cite{athalye2018obfuscated}, which can be used to attack non-differential prepossessing-based defenses. We used our defense technique that works against FGSM and BIM attacks to generate BPDA attacks. As other researchers \cite{carlini2019evaluating} pointed out that testing a defense in one dataset is not enough, therefore we chose multiple datasets (i.e., MNIST, CIFAR-10, and ImageNet). We considered a standard distortion $\epsilon$ = 0.3 for MNIST and $\epsilon$ = 8/255 for CIFAR-10, as current state-of-the-art \cite{tramer2020adaptive} recommend. Thus, our threat model is a combination of gradient-based, gradient-free, and adaptive evasion based adversarial attacks on multiple datasets. These attacks studied in this work are a combination of Whitebox, BlackBox, targeted and non-targeted attacks. Also, the presented defense will able to defend against attacks that are completely unknown to the proposed defense scheme. Uesato el al.\cite{uesato2018adversarial} advised to consider obscurity of adversarial attack when considering the defenses, which we also discussed in section \ref{experimentsresults}.
\subsection{Our Hypothesis}
Adversarial examples are input data that get misclassified by a ML/AI system but not by a human subject. 
let ML model be $M$, non-adversarial input data $A$, $A$'s true class is $C_R$ and noise data $\epsilon$, now $A_x= A + \epsilon$, $A_x$ classified by $M$ as class $C_W$, where $( C_W \neq C_R $). But if in human eyes $A_x \approx A$ and $A_x$ classify as $C_R $, $A_x$ is an adversarial example. $\epsilon$ is the added noise. 
\begin{figure}
\centering
  \includegraphics[scale=0.55]{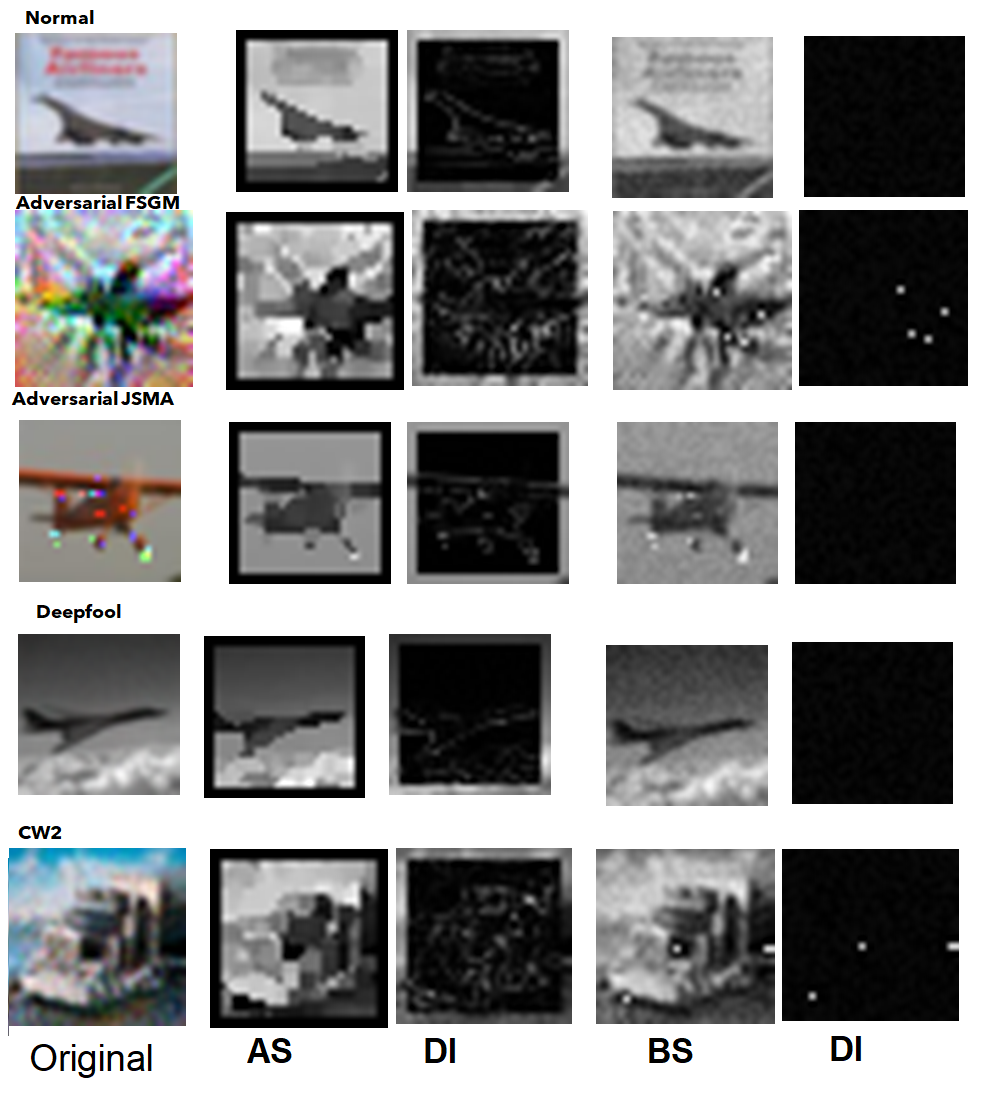}
  \caption{Effects of Several IPT techniques on CIFAR dataset.after applying different IPT, DI was done from grey-scaled image}
  \label{fig:cifar}
\end{figure}
Adversarial samples are distorted versions of non-adversarial samples. This distortion can measure in euclidean distances or pixel difference distance. However, we assume some image processing algorithms can signify the misuse of an adversarial example in a way that significance can be used to filter out adversarial examples. As an example, if an adversarial image has added perturb pixel all around its main edges, an edge-preserving technique algorithm can remove these perturb. So differences after-before image have a significant value which makes it distinguishable than a non-adversarial picture.

Let, Adversarial image sets are $A_{s}+\epsilon _s$, and clean image set is $A_{s}$, and IPTS denoted by $F_{ipts}$, $\epsilon_s$ is total added perturb of all adversarial images
So, after applied IPTS on adversarial and clean image set we will get,\\
$F_{ipts}(A_{s}+ \epsilon _s) \approx A_{s} + \epsilon _s +K_a$ \\
$F_{ipts}(A_{s}) \approx A_{s}+K$\\
where $K, K_a$ is the approximate effect of IPTS in clean and adversarial image set
So our difference DI is\\
$ DI \approx |(A_{s}+\epsilon _s) - (A_{s}+\epsilon _s +K_a) )- (A_s -A_{s}-K)|$ \newline
$DI \approx A_{s}+\epsilon _s - A_{s}-\epsilon _s -K_a - A_s +A_{s}+K|$\newline
$DI \approx |K - K_a|$ 
We can see that in our $DI$ equation no image ($A$ or $A_s$) is present. our aim is analyze the effects not the core image/image content but pre-processing effects. Note: $K, K_a$ is the generalized approximate effect of IPTS in clean and adversarial image set,

In testing time, We will calculate the $k_i$ value for the testimage $i$ by applying $F_{ipts}$.Here $k_i$ is the generalize effects from applied IPTS. Now $i$ will be a adversarial image if $|K-k_i|>|k_a-k_i|$. 
\begin{figure*}
\centering
  \includegraphics[scale=0.5]{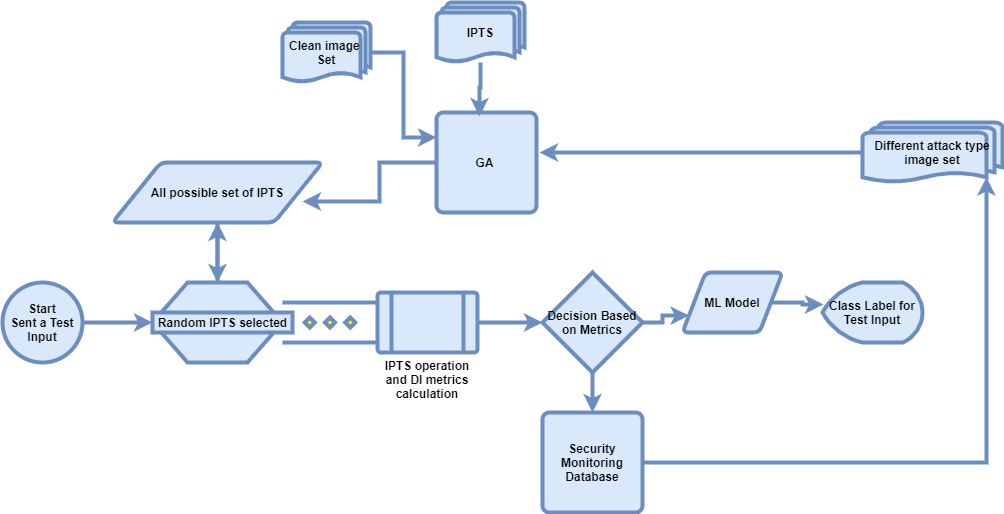}
  \caption{Illustrates the overall flow diagram  of our adversarial input detection technique. We used genetic search to find IPTS which can detect manipulation attacks on image datasets. During testing, a randomly picked SIPTS  will be selected for pre-processing the image, if the Image is not Adversarial it will go to ML system and get the result, otherwise, its information will be stored in a Security monitoring tool.}
  \label{fig:flow}
\end{figure*}
\subsection{Our Proposed Framework}
In this work, we developed a denoising approach to detect adversarial inputs using a sequence of image filters. This work inspired by Prakas et al.\cite{prakash2018deflecting}. But key differences are that our approach is automated and the out method does not need to consider if the attacks are Whitebox, Blackbox, targeted, non-targeted, or how the attack samples have been generated. Therefore, our approach is independent of the attack type, frequency, or models. Our method determines a suitable IPT sequence, which can detect adversarial image different attack types. From a cyber-security point of view, it is not enough to detect an attack. It is also necessary to know the attack type. This knowledge could help later on to assess the attacker's resources and help further to optimize defense. Adversarial samples are distorted versions of non-adversarial samples, such distortions can be measure in Euclidean distances or pixel difference distance. We assume that some IPT can signify the distortions of an adversarial example in a way that significance can be used to filter out adversarial examples. For instance, if an adversarial image has added perturbed pixels all around its main edges, an edge-preserving technique can remove these perturbations. The differences before and after processed images can be measured by different metrics (E.g: Histogram difference). These metrics will have significantly different values from the clean dataset's same metrics. We can differentiate between non-adversarial and adversarial images using these metrics. It could be possible that one unique sequence of edge-preserving and other IPT could make it more distinguishable than other sequences, thereby providing metrics of such differences to be used as a threshold value to classify adversarial and non-adversarial images. As there are a huge number of IPTS that exist, we use GA to find out one of the most appropriate ones. After that we will made set of IPTS (SIPTS) which cover all attack types and select them dynamically in testing time for detection.

\subsection{Problem Formulation}
Need to determine a set of IPT  $f(..)$, which can generate an optimal value of $\bar{\delta_h}$ and  $\sigma_{\delta_h}$ for a set of adversarial images $A_{x_(i=0...n)}$ and a non-adversarial set of image $A_{y_(i=0...n)}$ that we can use them to distinguish from non-adversarial image. Let IPT  be $f_1, f_2...f_n$. Same algorithm can be applied multiple times. The total number of algorithms applied for a set $F$ be $T$, where $T = count(f_1) + count(f_2) ... +count(f_n)$. For example, three IPTs are $f1, f2, f3$ and an optimal set for $F(f2,f3,f1,f1,f2,f1)$, here $T$ is $6$. So, our input is adversarial $A_{x}$ and non-adversarial $A_{y}$ set, and our output will be $f(..)$. and threshold value of different difference metrics $T_h$ (e.g., histogram average). When a test input will come, we will run $f(..)$ and determine the $T_{h_t}$.  If $T_h>T_{h_t}$ then it is an adversarial, otherwise it is a clean example.
\begin{table}[]
\centering
\begin{tabular}{|l|l|l|l|l|l|}
\hline
\multicolumn{1}{|c|}{\textbf{IPT}} & \multicolumn{1}{c|}{\textbf{2bit}} & \multicolumn{1}{c|}{\textbf{3bit}} & \multicolumn{1}{c|}{\textbf{IPT}} & \multicolumn{1}{c|}{\textbf{2bit}} & \multicolumn{1}{c|}{\textbf{3bit}} \\ \hline
Adaptive Smooth & 01 & 001 & Thining & - & 100 \\ \hline
Bilateral Smooth & 10 & 010 & Pixellete & - & 101 \\ \hline
Additive noise & 11 & 011 & Blur & - & 110 \\ \hline
Do-Nothing & 00 & 000 & Sharpen & - & 111 \\ \hline
\end{tabular}
\caption{GA Encoding: 2bit for MNIST, 3 bit for CIFAR \& Imagenet.}
\label{tab:gae}
\end{table}

\subsection{Selection of IPT}
There are numerous IPT that exist. Each can perform different operations in images. So, for a specific type of attack, a specific type of IPT sequence will work better. From our empirical observation, we selected a diverse set of unique IPT's such as BS, thinning \cite{goyal2011morphological}, AN, blur \cite{waltz1998efficient}, sharpen \cite{gross1998application}, thickening \cite{goyal2011morphological}, and AS. These techniques and their combinations are best suited to maximize enhancement between the difference image (DI)'s of adversarial and clean input. It can be observed in Figure \ref{fig:image} that, for the MNIST dataset, FGSM method tends to add pixelate noise around the object in the image, while JSMA seems to add along with the object borders, and CW tends to lose some erosion, whereas DF adds a very small amount of perturbing. Therefore, we assumed that sharpening can help to highlight JSMA attacks, performing blur will have more effect in CW, and AS will have a good effect on FGSM attacks. In figure \ref{fig:imageknn}, we can see different IPTS can work better for different attack types. So, we selected these IPTS, and found a combination of these IPTS that performs better. If we consider different datasets (e.g., ImageNet, CFIAR), we have to consider different IPTS, thus we select variety sets of IPTS, which can perform required all basic operations. The GA helps us to find a suitable set of IPTS. If we do not encode an IPT in the GA population, it will not be presented in any possible solution of the IPT sequence. Hence, the selection of IPT to generate the initial population is very important.
\subsection{Selection of IPT sequence (IPTS) using Evolutionary algorithm}
We will use a GA to solve our problem. The GA can do a heuristic search to find out an optimal set of IPTS. We used a steady-state GA. First, we will create our individuals from the set of filters. We will denote every IPT with binary digits, and these will work as a chromosome. If the encoding of IPTs are $F1, F2,..., F3$ then encoding can be seen as 
     F1=000,
     F2=001,
     F3=010,
     F4=011.
In practices, they encoded as table \ref{tab:gae}. Now, we will randomly generate the populations from the combinations of the IPTs. Here $I1, I2,...,In$ are individuals with total population $N$. Lets encoded them as
     I1= F2F1F2F4F2..Fi = 01010010101..1,
     I2= F1F2F2F4F3..Fi = 11010010101..1,
     I3= F2F2F1F4F3..Fi = 01011010101..1,
     I4= F1F2F2F4F4..Fi = 11010110101..1.
We will run I1, I2,…, In of IPT on $Ax$ and $Ay$ and we will get $DI_x$ and $DI_y$.  From $DI_x$ and $DI_y$ we will get  histogram average of $\bar{\delta_h}(x)$ , $\sigma_{\delta_h}(x)$ , $\bar{\delta_h}(y)$ , $\sigma_{\delta_h}(y)$ as per as formulation 1 for each individual.
Fitness value for histogram $F_{H}$ can be calculate as:
\begin{equation}
    F_{f_x}=|(\bar{\delta_h}(x_{I_i}) + \sigma_{\delta_h}(x_{I_i})) - (\sigma_{\delta_h}(x_{I_i}) + \bar{\delta_h}(x_{I_i}))|\end{equation}\begin{equation}
    F_{f_y}=|(\bar{\delta_h}(y_{I_i}) - \sigma_{\delta_h}(y_{I_i})) - (\sigma_{\delta_h}(y_{I_i}) - \bar{\delta_h}(y_{I_i}))|\end{equation}\begin{equation}
    F_{h} = |F_{f_x}-F_{f_y}|
\end{equation}

The following are the other four measures we implemented to measure the distances between adversarial and clean DI's histograms.
We measure cross-entropy as another fitness value $F_{CE}$
\begin{equation}
   F_{CE}  = \sum_{x} DI_{x_{hist}} (i) \log\bigg(\frac{1}{DI_{y_{hist}} (i)}\bigg)
\end{equation}
We also measure probability density (PD) cross entropy of DI histograms as another fitness value $F_{CE_{PDE}}$.
\begin{equation}
PD(DI_{x_{hist}})_{(i)}=\frac{1}{\sigma\sqrt{2\pi}}\exp\left[-\frac{1}{2}\left( \frac{i-\mu}{\sigma}\right)^2\right]
\end{equation}
\begin{equation}
F_{CE_{PD}}=  \sum_{x} PD(DI_{x_{hist}}) (i) \log\bigg(\frac{1}{PD(DI_{y_{hist}}) (i)}\bigg)
\end{equation}
We also calculate loss function to get another fitness value $F_{L}$.  
\begin{equation}
F_L=  \sum_{i=0}^n{-({Y\log(DI_{x_{hist}} (i))+(1-Y)log(1-DI_{x_{hist}} (i))}  ) }
\end{equation}
here $Y$ (= 0 and 1), respectively, denotes adversarial or clean sample. Using equation 4, 5, 6, 7 and 8, we get our fitness value. We normalize these values between 0 to 1. We weighted them based on their distance value compared to each other.
\begin{equation}
W_1=  \frac{F_{H}}{F_{CE}+F_{ED}+F_{CE_{PD}}+\frac{1}{F_{L}}}
\end{equation}
\begin{equation}
W_2=  \frac{F_{CE}}{F_{H}+F_{ED}+F_{CE_{PD}}+\frac{1}{F_{L}}}
\end{equation}
\begin{equation}
W_3=  \frac{F_{ED}}{F_{CE}+F_{H}+F_{CE_{PD}}+\frac{1}{F_{L}}}
\end{equation}
\begin{equation}
W_4=  \frac{F_{CE_{PD}}}{F_{CE}+F_{ED}+F_{H}+\frac{1}{F_{L}}}
\end{equation}
\begin{equation}
W_5=  \frac{\frac{1}{F_{L}}}{F_{CE}+F_{ED}+F_{CE_{PD}}+F_{H}}
\end{equation}
than we calculate our final fitness value $F_V$.
\begin{equation}
F_V=  W_1\times F_{h}+W_2\times F_{CE} +W_3\times F_{ED}+W_4 
\times F_{CE_{PD}}+W_5\times{\frac{1}{F_{L}}}
\end{equation}

We calculated $W$ values based on accuracy on detection from single fitness tests.
We will now sort the population based on the value obtained from each individual's fitness function. As it is a steady GA, we will do two-point crossover \cite{lin1997analysing} and replace the individuals with new off-springs. We will do elitism\cite{romero2016elitism} and keep some of the best individuals in our population. To avoid local optima, we will do mutation after selection. After the fitness values of the population converge, we will select the best individual for the IPTS for $A_x$ attack type. In the testing phase, we will check for the range of histogram, Euclidean distances, cross-entropy similarity, etc. to measure the input sample’s likelihood to be an adversarial attack or not.
\begin{figure*}
\centering
  \includegraphics[scale=0.3]{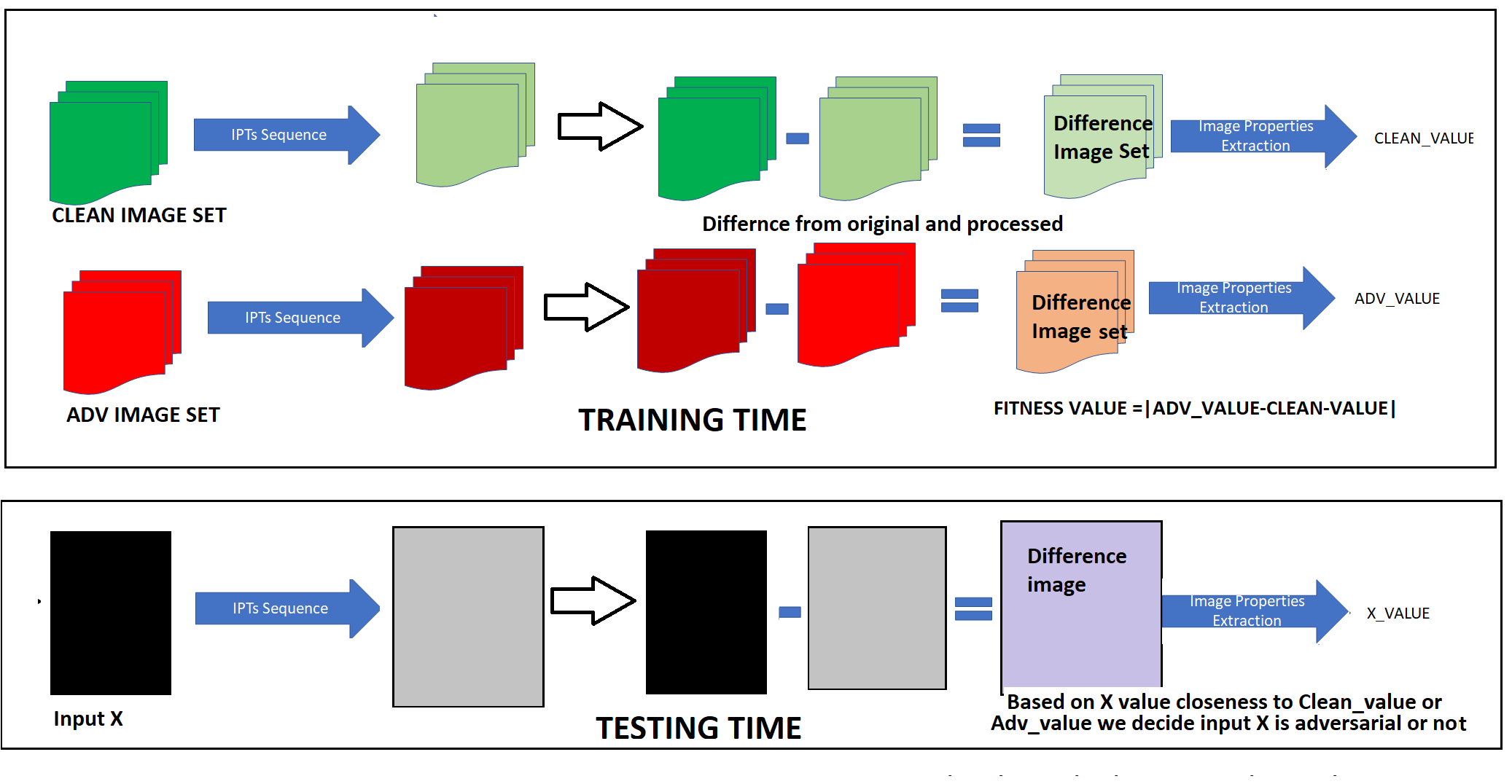}
  \caption{Sequence generation in training and testing time. A set of adversarial and clean processed using same IPT sequence and provide different value, distance of these values used in testing time. We used GA to determine best possible IPTS. So these values also used as fitness value in GA.}
  \label{fig:seqga}
\end{figure*}

\begin{figure*}
\centering
  \includegraphics[scale=0.6]{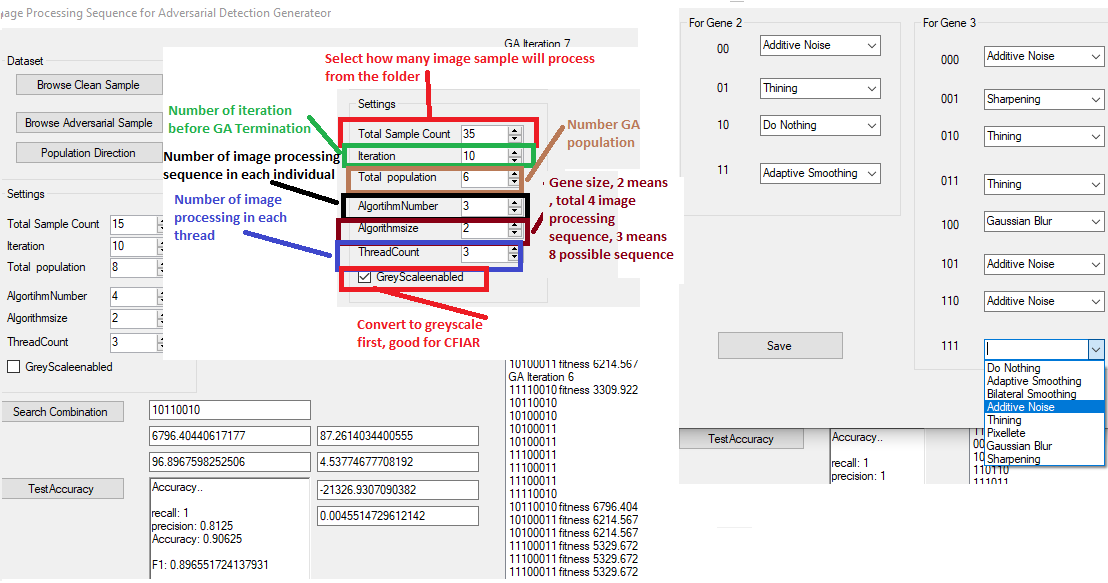}
  \caption{Our Demo UI for GA, Support 7 IPT.}
  \label{fig:exampleui}
\end{figure*}
\begin{table*}[]
\centering
\begin{tabular}{|l|l|l|l|l|l|}
\hline
\textbf{Dataset} & \textbf{Attack Type} & \textbf{IPTS} & \textbf{IPTS(decoded)} & \textbf{F1} & \textbf{\begin{tabular}[c]{@{}l@{}}GA \%\end{tabular}} \\ \hline
MNIST & PGD \cite{madry2017towards} & 111100 & Greyscale+2X Additive noise & 0.94 & 0.7 \\ \hline
CIFAR & PGD \cite{madry2017towards}& 111111101001 & 2X Sharpen+Blur+AS & 0.92 & 0.5 \\ \hline
ImageNet & FGSM \cite{goodfellow2014explaining}& 111111101001 & 2X Sharpen+Blur+AS & 0.86 & 0.3 \\ \hline
MNIST & JSMA \cite{papernot2016distillation} & 111111 & GreyScale+3X AN & 0.99 & 0.8 \\ \hline
CIFAR & JSMA\cite{papernot2016distillation} & 111001111000 & Greyscale+sharpen+AS+sharpen & 0.95 & 0.3 \\ \hline
ImageNet & JSMA \cite{papernot2016distillation}& 111001111000 & 2X sharpen+AS+sharpen & 0.79 & 0.6 \\ \hline
MNIST & CW \cite{xu2017feature}& 011011 & AS+BS+additive noise & 0.92 & 0.4 \\ \hline
CIFAR & CW \cite{xu2017feature}& 111111011101 & Greyscale+2x Sharpen+AN+blur & 0.94 & 0.3 \\ \hline
MNIST & DF \cite{moosavi2016deepfool} & 111101 & 2XAN +AS & 0.96 & 0.3 \\ \hline
CIFAR & DF \cite{moosavi2016deepfool} & 101101111011 & Greyscale+2blur+sharpen+AN & 0.95 & 0.2 \\ \hline
MNIST & HopSkipJump \cite{chen2019hopskipjumpattack}  & 101101 & BS+AN +AS & 0.97 & 0.6 \\ \hline
CIFAR & Localsearch\cite{narodytska2016simple} & 111101111111 & pixellete+ blur+pixellete+pixellete & 0.75 & 0.2 \\ \hline
MNIST & BPDA(PGD) \cite{carlini2019evaluating} & N/A & blur + pixellete +AS & 0.91 & N/A \\ \hline
CIFAR & BPDA(PGD) \cite{carlini2019evaluating}& N/A & blur + pixellete +AS & 0.89 & N/A \\ \hline
\end{tabular}
\caption{Different sequence generations for different attack types and datasets with their F1 score. GA \% is the success rate of the corresponding sequence selected as a solution in GA. For BPDA we tested against "blur+pixellate+AS" in PGD attack. This sequence works with 0.93 and 0.91 f1 scores against the PGD attack. With BPDA it gets reduces as shown in the table. Grey-scale is not encoded, it was predefined to speedup the process.}
\label{tab:resultsmy}
\end{table*}
\begin{table}[]
\centering
\begin{tabular}{|l|l|l|l|}
\hline
\textbf{\begin{tabular}[c]{@{}l@{}}Attack\\ Method\end{tabular}}  & \textbf{\begin{tabular}[c]{@{}l@{}}F1 (MNIST)\end{tabular}}&\textbf{\begin{tabular}[c]{@{}l@{}}F1 (CIFAR) \end{tabular}} \\ \hline
 PGD  & 0.91 & 089 \\ \hline
BIM  & 0.90 & 0.81\\ \hline
MBIM  & 0.91 & 0.88 \\ \hline
FGSM  & 0.98 & 0.97 \\ \hline
JSMA & 0.88 & 0.76\\ \hline
Hop-skip-Jump & 0.78 & NP\\ \hline
\end{tabular}
\caption{Experimental results using adaptive attack BPDA \cite{athalye2018obfuscated} (with defense sequence blur+AS+pixellete). }
\label{tab:resultsadaptive}
\end{table}
\begin{table*}[]
\centering
\begin{tabular}{|l|l|l|l|l|l|l|l|l|l|}
\hline
\textbf{\begin{tabular}[c]{@{}l@{}}Detection \\ Method\end{tabular}}                      & \multicolumn{4}{c|}{\textbf{MNIST}}                                                                           & \multicolumn{4}{c|}{\textbf{CIFAR}}                                                                          & \textbf{Avg}  \\ \hline
                                                                                       & \textbf{FGSM} & \textbf{JSMA} & \textbf{\begin{tabular}[c]{@{}l@{}}DF\end{tabular}} & \textbf{CW}   & \textbf{FGSM} & \textbf{JSMA} & \textbf{\begin{tabular}[c]{@{}l@{}}DF\end{tabular}} & \textbf{CW}   &               \\ \hline
\begin{tabular}[c]{@{}l@{}}Random Forest based adversarial training\cite{hayes2017machine}\end{tabular}                                & 0.96          & 0.84          & 0.98                                                          &   0.66            & 0.64          & 0.63          & 0.60                                                         &      0.72       & 0.77          \\ \hline
K-Nearest Neighbour based learning   \cite{hayes2017machine}                                                                                  & 0.98          & 0.80          & 0.98                                                          &      0.6         & 0.56          & 0.52          & 0.52                                                         &       0.69       & 0.73          \\ \hline
Supoort vector machine based learning  \cite{hayes2017machine}                                                                                  & 0.98          & 0.89          & 0.98                                                          &      -         & 0.69          & 0.69          & 0.64                                                         &        0.77      & 0.81          \\ \hline
\begin{tabular}[c]{@{}l@{}}Feature Squeezing\cite{xu2017feature}\end{tabular}                           & \textbf{1.00} & \textbf{1.00} &                                                               &      -         & 0.20          & 0.88          & 0.77                                                         &      -         & 0.77          \\ \hline
Ensemble \cite{bagnall2017training}                                                                               & 0.99          &      -         & 0.45                                                          &        -       & 0.99          &        -       & 0.42                                                         &      -         & 0.71          \\ \hline
\begin{tabular}[c]{@{}l@{}}Decision Mismatch\cite{monteiro2018generalizable}\end{tabular}                           & 0.93          & 0.93          & 0.91                                                          &  -             & 0.93          & \textbf{0.97} & 0.91                                                         &      -         & 0.93 \\ \hline
\begin{tabular}[c]{@{}l@{}}Image quality features \cite{akhtar2018adversarial}\end{tabular}                          & \textbf{1.00} & 0.90          & \textbf{1.00}                                                 &    -           & 0.72          & 0.70          & 0.68                                                         &        -       & 0.83          \\  \hline
\begin{tabular}[c]{@{}l@{}}Our Proposed  Method\end{tabular}                         & 0.99          & 0.99          & 0.96                                                          & \textbf{0.92} & 0.94          & 0.95          & \textbf{0.96}                                                & \textbf{0.94} & \textbf{0.96} \\ \hline
\end{tabular}
\caption{Comparison with other adversarial input detection technique based on F1-score.}
\label{tab:detection techniques}
\end{table*}

\begin{figure}
\centering
  \includegraphics[scale=0.3]{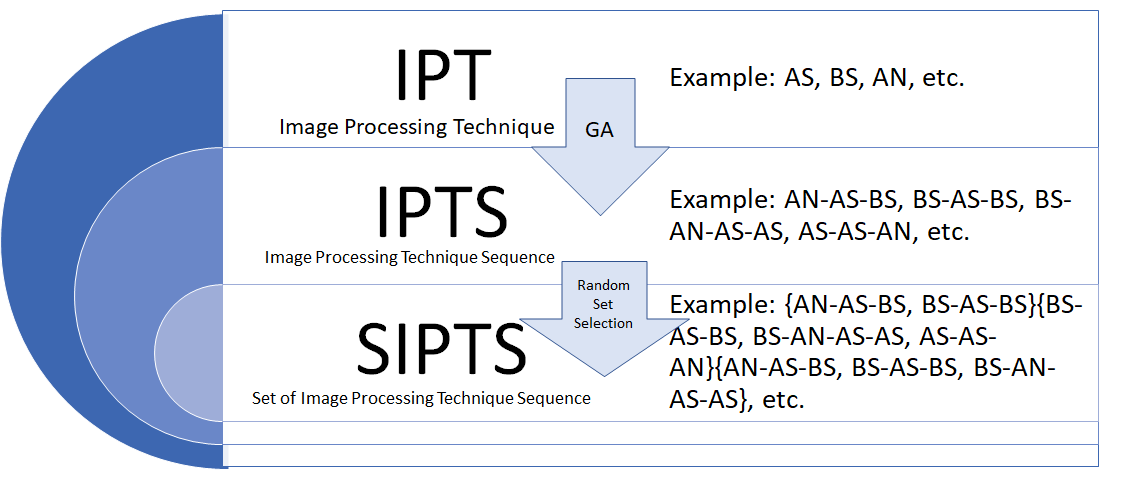}
  \caption{Illustrates the steps to generate SIPTS from IPT. At first, using GA we determine appropriate IPTS, As IPTS and attack types have many to many relations, Randomly choose one of the set which effective against all attack types. }
  \label{fig:explan}
\end{figure}
\begin{figure*}[hbt!]
\centering
  \includegraphics[scale=0.6]{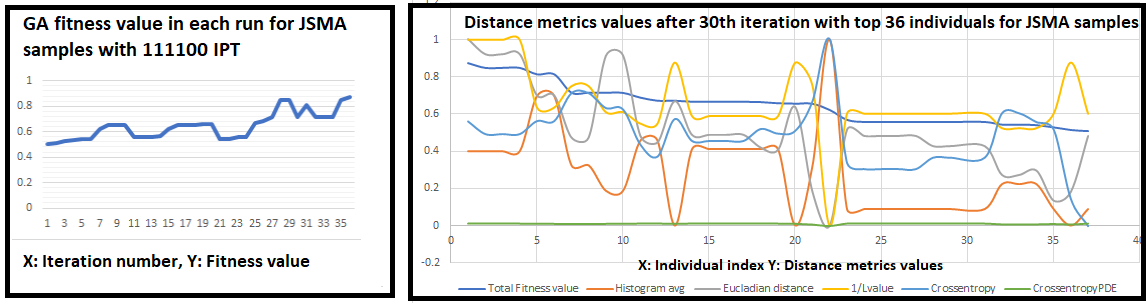}
  \caption{Here in the left side, we illustrated how GA best fitness value changes over each iteration and on the right side different metrics after 30 iterations for 36 individuals (IPTS)}
  \label{fig:gaana}
\end{figure*}
\subsection{Genetic algorithm implementation}
We need to find a sequence of IPT, which enhances the difference between adversarial and clean image data set, shown in Algorithm \ref{algo}. 
\begin{algorithm}
\caption{Genetic algorithm: Input: Adversarial image set $A_X$ and non-adversarial image set $A_y$}
\begin{algorithmic}
 \STATE \textbf{Population generation}: Create binary number sequence representing IPTs.
 \WHILE{Generation number == Termination number}
  \STATE \textbf{Selection Step 1}: Run each sequences $S$ from the population on $A_X$ and $A_Y$ and get $A_{X_S}$ and $A_{Y_S}$
  \STATE \textbf{Selection Step 2}: Generate difference image $A_{X_D}  =|A_X -A_{X_S}|$ and $A_{Y_D}  =|A_Y -A_{Y_S}|$
 \STATE \textbf{Selection Step 3}: Calculate image difference metrics of $A_{X_D}$  and $A_{Y_D}$, which represented as $A_{X_H}$ and $A_{Y_H}$ 
  \STATE \textbf{Selection Step 4}: Calculate fitness  value $F_V$ using equation 13 and Sort all sequence based on fitness values.
  \STATE \textbf{Crossover}: Remove lower fitness value individuals and do 2 point crossover
  \STATE \textbf{Mutation}:
  \ENDWHILE
  \STATE 1st individual in the population is output sequence

\end{algorithmic}
\label{algo}
\end{algorithm}
For input we need 
one set of adversarial images,
one set of clean images. 
N number of IPT represented by f1, f2, f3, … , fn. and in output GA will give 
a IPTS, e.g.,  “$f2
\rightarrow f3\rightarrow f2\rightarrow f1$” or
“$f3 \rightarrow f1 \rightarrow f2 \rightarrow f2$” . Based on N, we encode f1, f2,...,fn in binary sequence. If N = 4, the length L of binary representation will be 2, for N = 8, length L would be 3. For instance, for N=3 
f0=000,
f1=001,
..,
..,
F7=111.
We generate population with individuals with size\% L=0. So, if L is 2, size of individual can be 2, 4, 6. If L is 3, size of individual can be 3, 6, 9,…, etc.
Example:
$f7\rightarrow f0 \rightarrow f3 \rightarrow f1 \rightarrow f4 = 111 000 011 001 100$
We generate heuristic populations based on our insight of what IPT are more applicable for adversarial dataset.
Example: 
I1= 011100110011001,
I2= 101101010010111,
...,
I9= 000011001000011. 
Apply individual sequence on each image from both set and obtain distance metrics value from original and after application of sequence.
As example, For an individual $I1=111 000 011 001 100 = f7\rightarrow f0 \rightarrow f3 \rightarrow f1 \rightarrow f4$. If there is K number of images in an adversarial set, for each images in adversarial sample set, we applied $f7\rightarrow f0\rightarrow f3\rightarrow f1 \rightarrow f4$. If a image is $A$ and after of $f7\rightarrow f0\rightarrow f3\rightarrow f1 \rightarrow f4$ applied, it is $A^`$ then we calculate $A_d={A- A^`}$ and histogram value of $A_d$ is $H$. For K number of images, we will get $H_k$ number of histogram value. 
So average histogram value for adversarial set  will be $H_a=  \frac{H_k}{k}$.
Similarly, for M number of images in clean sample set, we will get average histogram value  $H_c =\frac{H_m}{M}$. Now, fitness value $F_{h}= (|H_a-H_c|)^2$     [we used squared value to increase the difference fitness value between each sequence]. Similarly, we will calculate all other fitness value, and using equation 13, we will get $F_{V}$. Now, for each individual, we will get different fitness value, e.g., I1= 011100110011001=fv1, I2= 101101010010111=fv2,...,I9= 000011001000011=fv9. We will sort based on fitness values,  will keep first sorted half, and do crossover between first half individuals to regenerate population. For mutation, we will flip a bit based on a probability of 0.05. After T iterations, we terminate the GA and pick the best individual in the population as our output sequence.

\begin{table}[]
\centering
\begin{tabular}{|c|c|c|c|}
\hline
\textbf{Attack type} & \textbf{MNIST} & \textbf{\begin{tabular}[c]{@{}c@{}} CIFAR\end{tabular}} & \textbf{\begin{tabular}[c]{@{}c@{}} ImageNet\end{tabular}}\\ \hline
FGSM & 0.98 & 0.97  & 0.86\\ \hline
Basic Iterative Method & 0.93 & 0.92 & NP \\ \hline
PGD(random start) & 0.95 & 0.98  & NP\\ \hline
MBIM & 0.99 & 0.98  & NP\\ \hline
hopskipjump & 0.70 & NP  & NP \\ \hline
JSMA & 0.69 &0.56 & 0.52 \\ \hline
CW & 0.36 & 0.28  & NP\\ \hline
DF & 0.41 & 0.32  & NP\\ \hline
\end{tabular}
\caption{IPTS generated using FGSM methods, applied on other attack types can detect with similar F1-score for variations of BIM attack types but some didn't perform well as they were not similar type of attack, NP:Not-performed}
\label{tab:perform}
\end{table}

\begin{table}[]
\centering
\begin{tabular}{|c|c|c|}
\hline
\textbf{\begin{tabular}[c]{@{}c@{}}Defense  Method\end{tabular}} & \textbf{DrawBacks} & \textbf{Advantages in Our Method} \\ \hline
\begin{tabular}[c]{@{}c@{}}JPEG compression, soft threshold\\ techniques, reduce the precision of data,\end{tabular} & \begin{tabular}[c]{@{}c@{}}Cant detect adversarial sample only \\reduces the probability of \\ miss-classification also static\end{tabular} & \begin{tabular}[c]{@{}c@{}}We can detect the attempt of\\ adversarial attack\\also we are dynamic\end{tabular} \\ \hline
\begin{tabular}[c]{@{}c@{}}Adversarial training\end{tabular} & \begin{tabular}[c]{@{}c@{}}Hampers the effectiveness of  model \\ and needs update regularly\end{tabular} & No need to access the ML model \\ \hline
\begin{tabular}[c]{@{}c@{}}Defensive distillation\end{tabular} & \begin{tabular}[c]{@{}c@{}}Needs two model to compare  Probability\\ which is expensive \\also it depends on ML model\end{tabular} & \begin{tabular}[c]{@{}c@{}}we use raw data not ML model\end{tabular} \\ \hline
\begin{tabular}[c]{@{}c@{}}Ensemble methods\end{tabular} & \begin{tabular}[c]{@{}c@{}}Multiple technique slow the\\ process also not effective\\ due to attack variations\end{tabular} & \begin{tabular}[c]{@{}c@{}}Our method is faster as it \\is only one set of operations\end{tabular} \\ \hline
\end{tabular}
\caption{Comparison with other defense techniques}
\label{tab:def}
\end{table}

\subsection{Selection of SIPTS}
\begin{figure}[hbt!]
\centering
  \includegraphics[scale=0.8]{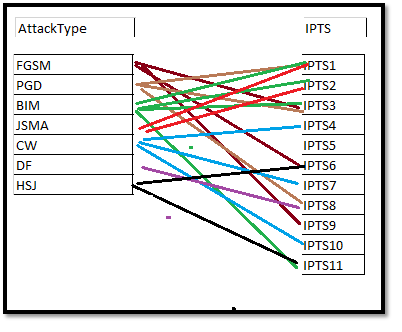}
  \caption{Many to Many relations between attack types and IPTS which works against these attacks}
  \label{fig:sipt}
\end{figure}
From each attack type, we selected 2-3 IPTS from the local optima of GA. From \ref{fig:sipt} We observed (a colored line from one attack type to multiple IPTS) that some IPTS also works for multiple attacks. There is multiple possible set of IPTS (SIPTS) exist which works against all attack types. For each test time, we will select different SIPTS for the test image. As example, we have evaluated 20 attack types and found 40 IPTS where 20 IPTS can resolve at least 2 attack types. We can have reasonably above 50k possible set of IPTS. If we randomly pick one for each test time, the chance of attacker to guess that one will be 1/50k which is close to 0.  So an attacker who has complete knowledge of our system and developed an adversarial sample that can bypass our IPTS (such as BPDA attack does) is very unlikely cause chance of guessing the IPTS is close to 0, our defense can work with 80\% accuracy against BPDA attacks, and making such samples are computationally high.

\subsection{Developed tool based on Proposed Framework}
We developed a desktop application, which can run a GA to determine the appropriate IPTS for each attack type. We named it `Image Processing Sequence for Adversarial Detection Generator Tool'. It takes adversarial image folder and non-adversarial image folder as input and generates image processing sequences as well as threshold value of histogram. In figure \ref{fig:exampleui}, a Basic GUI and configuration method of our developed tool is shown.
\section{Experiments}
\label{experimentsresults}
In this section, Details of how we generated attack samples and experiment details with one attack type (JSMA) are described.
\subsection{Target Dataset} 
Modified National Institute of Standards and Technology database (MNIST) consists of 60000 training and 10000 testing data of single digits zero to nine. These are all black background images with written digits in white colors. Using linear classifier it has 83\% accuracy. But with using CNN accuracy is above 99.6\%. Canadian Institute For Advanced Research (CIFAR) dataset includes 60,000 32x32 color images in 10 distinct classes (6,000 images of each class) . These classes are planes, cars, birds, deer, dogs, frogs, horses, cats,  ships, and trucks. The regular CNN has near 80\% accuracy, but ResNet has accuracy near 98\% and some efficient DNN methods reach the accuracy of 99\% \cite{wistuba2019survey}. ImageNet Dataset (IND) has more than 14million images, which have been handpicked and classified, it has more than 20000 categories such as ballon, strawberry etc. Using CNN made it possilbe to acheive 85\% accuracy here. Due to algorithmic bias, use of ImageNet is not always applicable as a standard dataset.

\subsection{Attack Samples generation} 
 For our initial testing, we applied AS, AN, BS, and thinning algorithms for MNIST. We added thinning, pixellete, blur and sharpen for CIFAR dataset. Our attack samples are from PGD, BIM, MBIM, FGSM, JSMA, DF, HopSkipJump, Localsearch, and CW methods. We generated minimum 250 adversarial samples from each attack type to ran our experiments. We find out a sequence of blur+AS+pixellete works for PGD, BIM, FGSM, MBIM. which we used as a defended model in BPDA adaptive attack and generated new attack samples. We generated FGSM, JSMA, CW, and DF using Pytorch \cite{gongzhitaao2018}. We used BIM, MBIM, PGD, local Search attack and HopSkipJump using IBM-ART-Toolbox\cite{art2018} and Cleverhans adversarial library\cite{papernot2018cleverhans}. We used modified advertorch\cite{ding2019advertorch} to generate BPDA attack samples. We noticed that the destruction rate (i.e., the rate of failure of adversarial attack when it is converted to visual form) \cite{kurakin2016adversarial} is higher in advanced attack types. We disregarded those images from attack samples. Also due to our restriction of  $\epsilon$ = 8/255 for CIFAR-10 as maximum noise value, we had to discard some samples from our dataset. For our GA, we used 75\% of random adversarial examples from our generated attack samples and the same number of random non-adversarial images from a clean data set. Remaining 25\% of adversarial images, combined with the same number of random non-adversarial images from the clean dataset is used for testing. Using the GA, we obtained an IPTS and difference metrics threshold values. When we applied that IPTS on the test set, we compare test data's difference metrics values with the threshold values. Based on that, we determine which images in the test set are adversarial and which ones not. We calculated true positive, false positive, true negative data and reported using F1 score.

\subsection{Experimental Setup}
We ran our experiment with different configurations to evaluate the impacts. In MNIST, we selected configuration as described in Table~\ref{tab:gae} for 2bit. Our initial individuals after 1st iteration are (every value multiplied with 100) 110110 fitness value 61.65, 000100 fitness value 51.11,...,110110 fitness value 00.92. After 20 iterations, we obtained population as 111100 fitness value 89.46, 111100 fitness value 89.46, ...,110100 fitness value 88.46. The best sequence was 111100 in this run, which decoded as two steps of additive noise. One of the distance metrics average histogram value has upper and lower range between 114-100. Using this range and other distance measures, we ran IPTS on 1000 image (500 are random FGSM samples and other 500 are random clean samples), and able to detect adversarial samples with F1 score: 0.98. For CIFAR data set, we found initial gray scale conversion improves the result slightly; we used configuration as we selected configuration as described in Table \ref{tab:gae} for 3bit. Our initial population of 30 individuals, after checking fitness (every value multiplied with 100) are 001001101110 fitness 21.17, 101111100010 fitness value 23.42,...,001010111100 fitness value 17.23. Our  population 30 individuals after 25th iteration are 111111101010 fitness 69.19, 111111101010 fitness 69.19,..., 111111101010 fitness 69.19. After more iteration we get best candidate $111111110010 \rightarrow 111-111-110-010$, which is decoded as $sharpen \rightarrow sharpen \rightarrow blur \rightarrow AS$, with range of 27-10. The obtained accuracy is 0.99 and F1 is 0.98. In figure \ref{fig:gaana} GA all iterations fitness values on the left graph and single iterations all individuals fitness values and distance metrics are shown in the right graph. Similarly we did experiments for all other attack types. After that we did a overall testing with all attack samples and choosing SIPTS randomly for each input. We achieved 97\% accuracy and 93\% F1-score with above 90\% precision and recall score.. 
\section{Result analysis}
\label{resultssectiona}
In the table, \ref{tab:resultsmy}, \ref{tab:resultsadaptive}, \ref{tab:detection techniques}, and , \ref{tab:perform} we presented our results in the F1 score for MNIST and CIFAR-10 dataset mainly. In this section, we analyze these results and describe how we are performing against the different aspects of our threat model.
\subsection{Defense against Common (gradient based) Attack}
In Table \ref{tab:resultsmy}, we observe that against common attacks such as PGD, JSMA, CW we are achieving over 90\% F1 score. It is also noted that the same IPTS also works for multiple methods. Here in the last column, we showed the percentage of time after 100 iterations we got the IPTS. There is a probability that a better IPTS exists if we iterate more. We did this experiment with three datasets. In MNIST we got the best result. For CFIAR we first convert an image to a GS image before applying the IPTS. In distance metrics calculation, we used the GS image as the original one. In IMAGENET we also converted to GS and then reduce the image size to reduce computation time, it may limit the reliability of our result for the IMAGENET dataset. 
\subsection{Defense Against Advanced Attacks}
When we were evaluating our defense against an advanced attack (with very low noises/perturbs) we found that as all adversarial attack types aim to reduce the perturbation in advanced attack types. The magnitude of perturbation gets so small that they get vanished in rounded values when  converting to visual form. Kurkin and Yan Goodfellow in their paper describe this phenomenon of destruction rate by the below equation \cite{kurakin2016adversarial}.
\begin{equation}
    d = \frac{\sum_{k=1}^n C(X^k,y{^k}_{True}) \overline{C(X^k_{adv},y{^k}_{True})} C(T(X^k_{adv}),y{^k}_{True}) }{\sum_{k=1}^n C(X^k,y{^k}_{True})C(X^k_{adv},y{^k}_{True})}
\end{equation}
Here, Destruction rate is the fraction of adversarial images which are no longer misclassified after the png or printed version conversion. 
Advanced attacks such as Deepfool, Local Search show a very high destruction rate which reduces our sample size as we exclude some samples from our experiment. In remaining samples, our test result is limited but if we combine the destruction rate with our detection rate the F1 score is above 90\%. As more advanced attack methods adversarial example generally cant have visualized form they are not effective as an adversarial input. As our method only work with adversarial inputs with visualized form, we could disregard those attacks. However, with 90\% accuracy, we can state that our defense is sufficient for low noise advanced attacks.

\subsection{Defense Against Attacks From Same Family}
In Table \ref{tab:resultsmy}, we observe that different sequences are required for different attack methods and different datasets. But, we also observe that the same IPTS is effective for all variations of BIM attack type. In Table \ref{tab:perform}, sequences generated using FGSM are working well with all the BIM variations. But, they completely failed when applied to CW or DF. Likewise, the IPTS which worked with CW but failed on FGSM attack methods.

\subsection{Defense Against Unknown Attacks}
Table \ref{tab:perform} proved that our method can defend against unknown attacks as BIM attacks were unknown to our defense. We also applied our defense selected for JSMA against attacks like CW and DeepFool and got 85\% accuracy. It proved that a set of IPTS will able to defend against unknown attacks when IPTS aren't determined uniquely. We ran an overall test against all attack types with single IPTS and got 80\% accuracy. When we run against a set of three IPTS we achieved 90\% accuracy. When we selected 6 IPTS, we achieved 98\% accuracy but the F1 score dropped to 90\%. An increasing number of IPTS in SIPTS increase the false-positive rate. So optimized SIPTS need to be selected for best results.

\subsection{Defense Against Adaptive Attacks}
An adaptive attack, such as BPDA \cite{uesato2018adversarial}, attackers bypass well known to preprocess techniques by applying a differential approximation. We developed a BPDA attack against one of the IPTS and we observed, other IPTS can easily detect the BPDA attack samples.
In table \ref{tab:resultsadaptive}, we can see that our defense technique performed well against adaptive attacks. Currently, we provided BPDA test results with only one IPTS. In the future, we plan to test it with other IPTS which are obtained using our GA. In the table \ref{tab:resultsadaptive}, we provided our results against BPDA attack. 

\subsection{Defense Considering obscurity}
About the obscurity as our defense sequences could be known to the attacker and they can devise new attack types that could bypass our defense technique. To address that, our GA can provide us multiple solutions, we will randomly pick one IPTS for one test input. The attacker won't know which IPTS he needs to bypass cause IPTS will be picked randomly. Also, our results against the BPDA attack supports that detection rate is higher even after the bypass. We also provided the probability of getting that sequence in each GA in Table \ref{tab:resultsmy}. This shows advanced attack type, optimal results have more chances to be different in each run.

\subsection{Adhering Cybersecurity Policies}
Our detection method is compliant with common security policies. All information security policy measures try to address three goals known as confidentiality (Protect the confidentiality of assets), integrity(Preserve the integrity of assets) and availability (Promote the availability of assets for authorized users). These goals form the  CIA model\cite{alexander2017cybersecurity} which is the basis of all security programs. Here, ML is considered as a digital asset, we ensure its confidentiality as we are not consuming or accessing any architectural information of ML. The same way integrity is preserved as we don't need to modify or tune anything in ML architectures. Many other defense techniques (example: \cite{papernot2016distillation}) access and modify the layers of deep learning models. Our method did not reduce the accuracy of ML like other techniques \cite{xu2017feature}, so 'Availability' is also preserved by our defense framework.

\section{Comparison with related works}
\label{resultssection}
We compared the proposed framework with existing adversarial attack detection methods and results are reported in Table \ref{tab:detection techniques}. We can see that on average our method has better accuracy.
For example, the feature squeezing based adversarial attack detection method developed by Xu et al.\cite{xu2017feature} attains average F1 score 0.77, whereas the proposed method in this study achieved 0.96 average F1 scores. Our method works for all types of ML model and does not need to evaluate any ML model's internal architectures. In this section, we provided a brief analysis of the comparison of our defense technique with other well-known techniques.
\subsection{Comparison with Adversarial training}
Adversarial training is adding adversarial examples in the training set and train the model. So the model can correctly classify adversarial examples. It is the most straightforward approach, but significant drawbacks for adversarial training is it reduces the accuracy of the ML model itself. Also, adversarial training can make the ML model more vulnerable to generalization \cite{raghunathan2019adversarial}. Aditi et. al. show that in CIFAR, adversarial training can reduce the accuracy from 95\% to 87\%. Also, some other drawback of this defense technique is we need to retrain the model whenever some new attack samples discovered. It will be hard to update all deployed ML models. 
For example, suppose some cars use the ML model to detect road signs. Every time some new adversarial method discovered full ML model need to retrain, evaluate and set up in the car again. Our method has some advantages over these drawbacks as we don't touch the actual ML model, so our model doesn't affect the accuracy of the model anyway. Also, in case of a new attack method discovered, we only need to update our filters threshold values, we don't need to update the model.

\subsection{Comparison with Image Prepossess techniques}
Most of the image processing techniques reduce the adversarial effect before sending it to the ML model. The major drawback of these techniques is their processing techniques are static; it doesn't evolve as alongside the attack. Also, these models don't detect an adversarial sample. But for security purposes, it is essential to know if an ML method is under attack or not. One of the image processing techniques by Prakash et. al. \cite{prakash2018deflecting} is detect adversarial samples, but it is a static approach. So it is very much vulnerable to zero-day attacks. Also, all of the related methods have a single sequence of image processing techniques that do not apply to all adversarial attack methods. Our plan overcome this by using a genetic algorithm to do a heuristic search of suitable courses to filter out adversarial samples.

\subsection{Comparison with Distillation technique}
Distillation techniques work with combining the double model, the second model uses the first model knowledge to improve accuracy. The recent advancement in the black-box attack make this defense obsolete \cite{chakraborty2018adversarial}. Strong transfer-potential of adversarial examples across neural network models \cite{papernot2016distillation} is the main reason for this techniques failure. Also, Carlini and Wagner proved that this method doesn't work against CW attack methods. It is not robust as simple variation in a neural network can make the system vulnerable to attacks \cite{carlini2016defensive}. The advantage of our method over defense distillation is it doesn't need to modify neural network and it also works against CW attack methods. 

\subsection{Comparison with MagNet}
Meng et al. \cite{meng2017magnet} proposed a framework, named MagNet, which uses classifier as a blackbox to read the output of the classifier’s last layer without reading the data on any internal layer or modifying the classifier. As it needs the information of ML last layer, thus it is violating the Confidentiality property of CIA (Confidentiality, Integrity, Availability) triad. Our proposed method doesn't need to read any layer of the ML model. So our model remains the same for both Blackbox and Whitebox attack methods. 

\subsection{Comparison with Ensemble techniques}
Combining multiple techniques to create a defense against the ML model is another approach some researchers took. But from the study of He et el. \cite{he2017adversarial}, it was concluded that combining weak defenses does not automatically improve the robustness of
a system. Developing effective defenses against adversarial examples is an important topic. Also, this technique remains static and vulnerable to a new attack. Our proposed solution doesn't have cause we select defense technique dynamically in testing phase.

\subsection{Comparison with Feature Squeezing}
Feature squeezing is another model hardening technique \cite{xu2017feature}. This method reduces the data in a way the perturbed value looses more in the process as perturbs have low sensitivity. For example reducing the image by 1 bit will not change the model outcome much but it will remove the adversarial perturbations. However it still affects the accuracy of ML model. In our proposed solution there is no such effect on actual model accuracy.

\subsection{Comparison with Defense GAN} Samangouei et. al. \cite{samangouei2018defense} proposed a mechanism to leverage the power of Generative Adversarial Networks to reduce the efficiency of adversarial perturbations. This method is very effective against white and blackbox attack. But efficiency of GAN depend om the training of GAN which is computationally complex and need suitable data-sets. Our method doesn't need complex training method.

\section{Conclusion and Future work}
\label{conclusion}
Limitations of our current implementation is that our GA-Based approach for detecting adversarial inputs, used a sequential computational platform, which took a long time to find the optimal IPTS. While the current work focused on the concept and feasibility of the proposed approach, we plan to use Hadoop/spark big-data clusters with a larger dataset in our future work. This work demonstrated that the proposed method can detect basic attacks, and it is also equally effective against variations of these attack types. Karkin et al.\cite{kurakin2016adversarial} and Lu et el\cite{lu2017no} showed that advanced adversarial attacks have less adverse effects when converted to a visual form (such as PNG). Hence, they have less chance to remain adversarial in the real world. Inspired by these works, we discarded some samples, (mostly from advanced attacks) which limits our total sample size. Because of space limitations, we could not provide complete results from all other attack types and their adaptive-ness with the BPDA attack here. Experimental results on three datasets showed that the presented method is capable of detecting various types of attacks, including adaptive attacks. Since this method does not need any knowledge about the main ML model whose input it is detecting, thus it can work as an independent input data filter for any ML models. Also, it can classify what type of attacks are coming and when such attacks are happening, which is valuable from a security perspective. Future work will further explore parameter tuning as well as variable-length representation, ordered tuples, specialized operations, adding more IPTS and, other difference measures, etc. We aim to implement this GA to work in cloud computing (Hadoop/Pyspark) to speedup the appropriate sequence estimation.

\bibliographystyle{plain}
\bibliography{main.bib}

\end{document}